\documentclass[conference,compsoc]{IEEEtran}
\IEEEoverridecommandlockouts

\ifCLASSOPTIONcompsoc
  \usepackage[nocompress]{cite}
\else
  \usepackage{cite}
\fi
\ifCLASSINFOpdf
   \usepackage[pdftex]{graphicx}
\else
\fi
\usepackage{amsmath}
\usepackage{CJKutf8}
\usepackage{multirow}
\begin{document}
\begin{CJK*}{UTF8}{gbsn}
%
\title{Crowdsourcing Argumentation Structures in Chinese Hotel Reviews}

\author{
\IEEEauthorblockN{Mengxue Li\IEEEauthorrefmark{2},
Shiqiang Geng\IEEEauthorrefmark{3}${*}$\thanks{${*}$The first two authors contributed equally to this work and should be considered co-first authors.},
Yang Gao\IEEEauthorrefmark{2},
Haijing Liu\IEEEauthorrefmark{2}
and Hao Wang\IEEEauthorrefmark{2}
}
\IEEEauthorblockA{\IEEEauthorrefmark{2}Institute of Software Chinese Academy of Sciences, University of Chinese Academy of Sciences\\
\IEEEauthorrefmark{3}School of Automation,
Beijing Information Science and Technology University, China\\
Email: mengxue2015@iscas.ac.cn, gsqwxh@163.com, \{gaoyang, haijing2015, wanghao\}@iscas.ac.cn }
}

%


\maketitle

\begin{abstract}
  \emph{Argumentation mining} aims at automatically 
  extracting the premises-claim discourse structures in 
  natural language texts. 
  There is a great demand for argumentation corpora
  for customer reviews. 
  However, due to the  controversial nature
  of the argumentation annotation task, 
  there exist very few large-scale argumentation
  corpora for customer reviews. 
  In this work, we novelly use the crowdsourcing technique
  to collect argumentation annotations in Chinese hotel reviews.
  As the first Chinese argumentation dataset,
  our corpus includes 4814 argument component
  annotations and 411 argument relation annotations,
  and its annotations qualities are comparable
  to some widely used argumentation corpora in other languages.
\end{abstract}


%
\IEEEpeerreviewmaketitle

\section{Introduction}
\label{sec:intro}


In customer reviews,
users usually not only give their opinions
on the products/services, but also provide reasons
supporting their opinions. For example, consider
the following review excerpt posted on Tripadvisor.com:

\begin{quote}
  \textbf{Example 1}:
  \textcircled{1} 房间的电器设施让人很失望。
  \textcircled{2} 有一台很老很小的黑白电视。
  \textcircled{3} 空调也是坏的。
  \textcircled{1} Appalling in room electrical facilities. 
  \textcircled{2} There was an old, small, black TV. 
  \textcircled{3} Air conditioner did not work.  
\end{quote}

Clause \textcircled{1} gives the customer's 
opinion (or \emph{claim}) on the appliances in the room,
and clauses \textcircled{2} and \textcircled{3} are 
reasons/evidences (or \emph{premises}) supporting
the claim. Such discourse structures are known as \emph{arguments}
\cite{moens2013argumentation},
and the techniques for automatically extracting
arguments and their relations (e.g. support/attack)
from natural language texts are known as 
\emph{argumentation mining} \cite{lippi2016survey}. 
Performing argumentation mining on customer reviews
can reveal the reasons behind users' opinions, 
thus can greatly facilitate the product producers and 
service providers to figure out their weaknesses
and hence has huge commercial potentials.

There exist a great demand for reliably annotated 
corpora on customer reviews, 
since they are required for training 
supervised-learning-based argumentation mining techniques.
Existing argumentation corpora are mostly constructed from 
highly professional genres, e.g. legal documents \cite{palau2009argumentation},
persuasive essays \cite{ig2016parsing}, newspapers and court cases \cite{reed2008corpus}.
Compared to these genres, 
customer reviews are written by ordinary people in casual scenarios, 
thus their linguistic complexities 
are usually lower and do not contain much domain knowledge;
as a result, we believe that even novice people are able to identify the 
argumentation structures in customer reviews.
Crowdsourcing has been widely recognised as a reliable 
and economic method for some annotating
tasks \cite{snow2008crowdsourcing}. In this work, we investigate the 
applicability of crowdsourcing for argumentation annotation
in Chinese hotel reviews.
Specifically, the contributions of this work are threefold:
\begin{itemize}
  \item 
    We propose a novel \emph{argumentation model}\footnote{An 
    argumentation model gives the definition of arguments, 
    e.g. what components an argument is consisting of, what kinds of
    relations are allowed between different argument and argument components.} 
    for hotel reviews, 
    which extends the classic ``premise-claim'' model, and can be
    potentially used for defining argumentation structures
    in other types of customer reviews and in other languages.
  \item
    We novelly employ crowdsourcing to annotate
    argumentation structures (i.e. argument components
    and their relations) in Chinese hotel reviews, 
    design some mechanisms so as to help the workers 
    reduce their chances of making mistakes, and use
    a clustering algorithm to aggregate collected annotations.
  \item
    The aggregated annotations are published as a publicly available
    corpus, and the annotating quality of the corpus
    is comparable to the state-of-the-art English argumentation
    corpora. Furthermore, because of the controversial nature of the 
    argumentation annotation task, 
    we provide a \emph{confidence score} to each label, 
    so as to help users understand the controversy degree
    of each annotation. To the best of our knowledge, 
    this is the first Chinese argumentation corpus,
    and the first use of confidence score in argumentation corpora.
\end{itemize}

\section{Related Work}
\label{sec:related_work}
We first review existing argumentation corpora
for customer reviews;
in particular, we highlight the argumentation
models they used to define arguments.
After that,
we review works on crowdsourcing for 
argumentation annotation and some related tasks,
e.g. annotating discourse structures.

\subsection{Argumentation Corpora}
\label{subsec:related:arg_corpora}
A comprehensive review on argumentation corpora
is beyond the scope of this paper;
good overviews can be found in e.g. 
\cite{lippi2016survey,ig2016web,ig2016parsing}.
Here we only review argumentation corpora constructed from
customer reviews. 

Wachsmuth et al. \cite{wachsmuth2014review} built the 
\emph{ArguAna} corpus, consisting of 2.1k hotel reviews
posted on Tripadvisor.com. Instead of directly
labelling arguments, they annotate \emph{statements}
and sentiments polarities.
A statement is ``\textit{at least a clause and at most a sentence 
that is meaningful on its own}''. 
They designed a rule-based tool to segment statements, 
and employed crowdsourcing to annotate
the sentiment and \emph{features} 
(e.g. location, services, facilities)
in each statement. 
Results suggested that crowdsourcing workers can reliably 
identify the sentiments of statements (approval rate 
72.8\%) but have controversies for identifying features
(rejection rate 43.3\%).
We view ArguAna as an intermediate resource 
for building argumentation corpora, because each 
argument usually contains several statements (e.g.
one positive/negative statement serving as the claim,
and several neutral statements serving as premises). 
Thus, ArguAna cannot be directly used for
training argumentation mining techniques. 

Garcia Villalba and Saint-Dizier \cite{villalba2012some} 
investigated suitable
argumentation models for customer reviews. 
They viewed  many different type of expressions
(e.g. illustrations, elaborations and reformulations) 
as argument components, 
and built a corpus consisting of 50 customer reviews
in French and English in the domains of hotels and restaurants, 
hifi products, and the French political campaign.
Wyner et al. \cite{wyner2012semi} built a corpus 
consisting of 84 reviews (posted on Amazon.com) 
for one specific camera model.
They considered one specific argumentation
model: an argument consists of two premises
(premise 1 gives ``camera X has property P'' and
premise 2 gives ``property P promotes value V'')
and one claim (the customer should perform action ACT;
possible ACT include ``buy the camera'',
``avoid using the flashlight'', etc.).
However, for both these corpora, their \emph{inter-rater agreement}
(IRA)\footnote{IRA is a widely used metric to 
evaluate the annotation quality. 
There exist multiple methods for computing 
the IRA score; in this paper, for each IRA score,
we will point out the computation method it uses.
Larger IRA values suggest higher agreement between
the annotators, thus suggest higher reliability 
of the obtained annotations.} were not reported,
and they were not publicly available.



\subsection{Crowdsourcing for Argumentation 
Annotation and Related Tasks}
\label{subsec:related:crowdsourcing}

Ghosh et al.
\cite{ghosh2014crowdsourcing} proposed an 
annotation mechanism to annotate arguments and their
relations in blog comments: they 
hired crowdsourcing workers to
label claims-premises relations. 
Note that the arguments segments were provided
a priori (annotated by domain experts), 
and the crowdsourcing workers
were asked to only label the argument 
component types and relations. They reported that
crowdsourcing workers achieved 0.45-0.55 IRA score
(in terms of multi-$\pi$ \cite{fleiss1971measuring}), 
suggesting the agreement is moderate; also, they suggested that
the agreement scores of the crowdsourcing workers 
is highly correlated to those of the expert annotators.

Crowdsourcing has been used to annotate discourse
structures. Kawahara et al. \cite{kawahara2014crowdsourcing} designed
a two-stage crowdsourcing mechanism to annotate 
two levels of discourse relations in 
Japanese texts crawled from multiple 
online genres. In their work, discourse relations
include contrast, concession, cause-effect, etc.,
and these relations closely resemble some
relations in argumentation structures (e.g. 
the attack relation between arguments can be 
viewed as contrast, and the premise-claim
relation is closely related to the cause-effect
relation). They did not report the IRA
of the crowdsourcing workers, but instead,
testified the quality of their discourse corpus
by training a discourse parser on their corpus.
Results suggested that the quality of their corpus is
comparable to the state-of-the-art English discourse
corpus, indicating that crowdsourcing can be reliably 
used for annotating relations between clauses.

\section{Pre-Study}
\label{sec:pre_study}
In this section, we describe how we design our annotation guideline.
In particular, we perform some preliminary annotating
experiments to decide
i) how to segment clauses (e.g. by rule-based automatic methods or 
by crowdsourcing workers), and ii) which argument model to use,
i.e. which argument components constitute an argument,
and what relations are permitted between argument components.
Since most existing annotation guidelines are for English texts,
we annotate twenty English and ten Chinese hotel reviews
(without titles) from Tripadvisor.com 
to draft our guideline. Five annotators participate
in the experiments; they are all Chinese
native speakers fluent in English.
The annotation is performed on the \emph{brat} \cite{stenetorp2012brat}
open-source annotation platform. 

As for clause segmenting, we test two approaches:
sub-sentence based segmenting \cite{wachsmuth2014review}
(i.e. viewing each sub-sentence as a clause), 
and free segmenting (i.e. any span of texts 
can be viewed as a clause). 
We find that the free segmenting strategy is more
suitable for Chinese hotel reviews, because punctuations
are often missing or misused.
In addition, the free segmenting strategy enables the annotators
to label the exact boundary of each argument component,
avoiding including some connecting words in argument
components (e.g. ``而且''(in addition)). 

As for the argumentation model, we consider three
candidate models: 
the Premises-Claim-MajorClaim (PCM) model \cite{ig2014coling} 
for persuasive essays, the extended Claim-Premises (ECP) model
\cite{habernal2014argumentation} for long 
Web documents and the 
extended Toulmin's (ETM) model  
\cite{habernal2014argumentation} for short Web documents.
Our experiments suggested that:

\begin{itemize}
  \item 
    In both Chinese and English hotel reviews, users
    often give their overall impression on 
    hotels (e.g. 强烈推荐(strongly recommend), 
    我觉得是很不错的酒店(I think the hotel is quite good),
    很美好的回忆(living here leaves me beautiful
    memories)). We believe the \emph{MajorClaim} 
    component in the PCM model is the most 
    suitable argument component type for labelling
    these clauses.
  \item
    In hotel reviews, users often give evidences supporting
    some implicit claims: consider the clause
    ``离市中心走路不到五分'' (just 5 minutes' walk
    to the city center),  we believe this clause is not only
    for stating the location of the hotel, but also supports some
    claims (e.g. the location of the hotel is good), although these
    claims are not explicitly presented in the hotel reviews. 
    We thus believe that the argument component \emph{premise supporting implicit
    claim} (PSIC) should be used in our argumentation model.
  \item
    Distinguishing different kinds of premises (e.g. grounds
    warrants, refutation) in hotel reviews is error-prone even
    for expert annotators; thus we decide not to use the ETM  
    model, although ETM is proposed for short user-generated documents.
\end{itemize}

\begin{figure}
\centering
\includegraphics[scale=0.3]{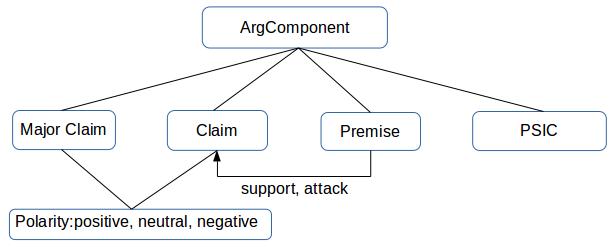}
\caption{The argumentation model we proposed for hotel reviews.}
\label{fig:arg_model}
\end{figure}

Our argumentation model is an extension and integration
of the PCM and ECP models; it includes four argument components:
MajorClaim, Claim, Premise and PSIC; 
texts not labelled as any argument component
are \emph{non-argumentative} (NA for short).
Premises are allowed to support/attack claims,
but claims are not allowed to 
support other claims, because this will lead to \emph{cascading support},
which overcomplicates the annotating process \cite{habernal2014argumentation}.
In addition, annotators are also asked to annotate the sentiment
polarity (positive, negative or neutral) for MajorClaims and Claims,
because these two argument components are subjective.
Fig. \ref{fig:arg_model} illustrates our argumentation model, 
and some examples are given in Table \ref{table:arg_model_example}. 
Note that each premise must support/attack some claim,
but a claim may have no premises supporting/attacking it. 

The final annotation guideline is in Chinese, 
including detailed explanations of the argument model,
segmenting rules as well as considerable illustrative examples.
To test the readability and applicability of our annotation
guideline, we ask additional two students (Chinese native speakers)
to independently annotate five Chinese hotel reviews
using our annotation guideline. The average IRA 
in terms of Krippendorff's $\alpha_U$ \cite{krippendorff2004measuring}
for their annotations is 0.715,
suggesting that the agreement is substantial.

\begin{table}
\caption{Examples of argumentation model we propose. 
Clauses are in square brackets, and argument component types
are in round brackets.}
\label{table:arg_model_example}
\centering
\footnotesize
\begin{tabular}{p{2cm}|p{5.5cm}}
\hline
{\bf Type} & {\bf Examples}
\\
\hline
Major Claim & ``总之，[这个酒店很好](Major Claim1)，[我很满意](Major Claim2)。'' 
``In short, [this hotel is very good] (Major Claim1), [I am very satisfied
with this hotel.] (Major Claim2).''
\\ 
\hline
Claim & ``[舒适的环境](Claim1)和[周到的服务](Claim2)，[房间设施也很齐全](Claim3)。''
``[Comfortable environment] (Claim1) and [thoughtful service](Claim2), 
and [the room has all necessary appliances and equipment](Claim3)."  
\\
\hline
\multirow{2}{*}{Premise} & 
\textcircled{1} ``[服务员很好](Claim)，[会主动帮我们拿行李](Premise)。''
``[The staff are very nice] (Claim), [they came out to help us take our luggage] (Premise)." \\
&
\textcircled{2}
``[酒店餐厅的美食还挺不错](Claim),就是[食物的价格都比较贵](premise)。''
``[The food in the dining hall of this hotel is pretty good] (Claim), 
despite [the steep price] (Premise)."
\\
\hline
PSIC & ``【房间的冰箱里面就有水，还有一些其它的吃的东西，都是可以先享受再付费这样的】(PSIC)。'' ``[There are water in room's refrigerator and some other food, you can pay after you use them] (PSIC)." 
\\ 
\hline
Support & In example \textcircled{1}, the Premise supports the Claim 
\\ 
\hline
Attack & In example \textcircled{2}, the Premise attacks the Claim 
\\ 
\hline
\end{tabular}
\end{table}

\section{Crowdsourcing Experiments}
\label{sec:crowdsourcing}
Our crowdsourcing experiment is performed as an optional
assignment in the \emph{social media mining} course
in University of Chinese Academy of Sciences in 2017. 
Over four-hundred MSc students are registered for this course,
above 90\% are Chinese native speakers.
We call for volunteers to participate in our experiment,
and inform them that the participated students can obtain
the corpus and its statistics in return, which they can use
in their final project to train some machine learning algorithms.
At last, 388 students participate and we give 
an one-hour tutorial to help them go through the guideline
and to illustrate some examples.

The crowdsourcing is performed on the \emph{brat} \cite{stenetorp2012brat}
platform. 
Each student can only view the hotel reviews
assigned to him/herself. 
To help the annotators reduce mistakes, 
we customise and extend the original brat system so that 
i) only legal relations (see Fig. \ref{fig:arg_model}) 
are allowed to annotate, 
ii) the annotator is reminded if 
the sentiments for some MajorClaims or Claims are not labelled,
and iii) the annotator is reminded if there exist some premises
that do not support/attack any claims (this is not allowed;
see Sect. \ref{sec:pre_study}).

Each student receives twenty-five hotel reviews to annotate;
one in these twenty-five reviews is a gold standard review,
which has been annotated by our expert annotators in our pre-study
and is used to evaluate the devotedness of the annotator.
Each non-gold-standard hotel review is allocated to 4 students 
to independently annotate. All hotel reviews are crawled from 
Tripadvisor.com.cn,
and their length are all between 100 to 150 words.
Students are asked to finish all labelling in one week.
At last, we collect 2332 hotel reviews' annotations.
Because around 10\% students (38 students) fail to annotate 
all document assigned to them, some documents receive only one 
student's annotation, and we remove these documents.
In addition, we remove the annotations that violate our
annotation guideline. In total, 7 hotel reviews
and their annotations are removed. 

\begin{figure}
\centering
\includegraphics[width=0.45\textwidth]{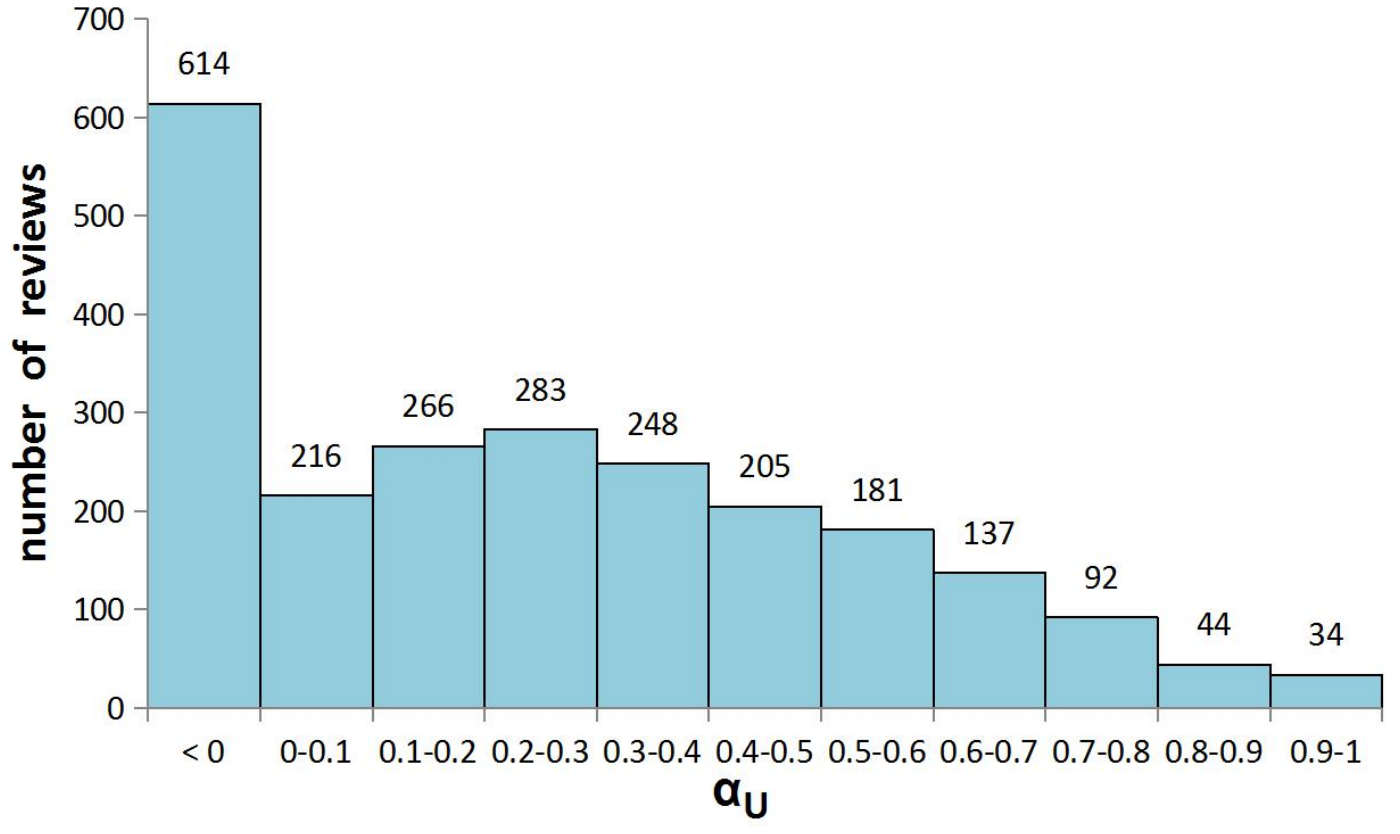}
\caption{The distribution of $\alpha_U$.}
\label{fig:raw_histogram}
\end{figure}

We compute $\alpha_U$ 
for each hotel review\footnote{
  At the moment, the $\alpha_U$ for a hotel review
  considers only the annotations for component types,
  and ignores the agreement for annotations on 
  sentiment and relations, because we observe that
  the agreement scores
  for sentiment and relations are highly correlated
  to the $\alpha_U$ for argument component. Detailed 
  agreement for sentiment and relations will be presented in
  Sect. \ref{subsec:post_process:controversial}.
} so as to evaluate the quality of annotations.
The distribution of $\alpha_U$ scores is presented in 
Fig. \ref{fig:raw_histogram}. We find that
only 21\% documents receive $\alpha_U \geq 0.5$,
suggesting that the annotations for most documents
are quite diverging and cannot be directly used 
to build the corpus. 
We believe that two factors may have resulted in
the low annotation agreement: 
i) the low devotedness of some annotators, and 
ii) the controversial nature of argumentation annotation
in some hotel reviews. 
In the next section, we will give a specific analysis of these two reasons and attempt to improve
the annotation quality along these two lines,
i.e. by removing less-devoted students' annotations
and by identifying controversial sentences, 
so as to produce a high-quality Chinese hotel review corpus.

\section{Post-Processing and Corpora Generation}
\label{sec:post_process}

We identify the less-devoted students
and remove their annotations in 
Sect. \ref{subsec:post_process:students},
compute the confidence score for annotations
and generate the final corpora in 
Sect. \ref{subsec:post_process:controversial},
and perform the error analysis in Sect. \ref{subsec:error_analysis}.

\subsection{Remove Less-Devoted Students' Annotations}
\label{subsec:post_process:students}

We use the gold standard annotations to evaluate
each student's devotedness.
The gold standard texts are similar
to the examples in the guideline, thus we believe that
the students whose annotations diverge widely from
the gold standard annotations are less devoted.
For each sentence in a gold standard text, 
we compute all students' agreement (in terms of 
$\alpha_U$)
against the gold standard annotation on this sentence,
and rank these agreement scores.
If a student's agreement score on this sentence 
falls into the bottom 10\%, we increase this student's
less-devotedness degree by 1.
Students whose less-devotedness degrees are equal or
larger than 2 are labelled as less-devoted,
and all of their annotations are removed.
In total, we found 38 less-devoted students, 
and we additionally remove 39 hotel reviews
because they receive fewer than two students'
annotations after deleting the less-devoted 
annotations.

We find that removing the
less-devoted students' annotations can indeed
improve the annotation quality as a whole.
For example, after removal, the percentage of 
reviews whose $\alpha_U \geq 0.5$ increases
from 21\% to 24\%.
In addition, the average agreement score for
each gold standard text has also be increased
thanks to the removal (see Table \ref{table:alpha_of_golddata}).
However, even after the removal, there still exist
considerable controversial annotations, due to the controversial
nature of our guideline and the argument annotation task itself.
Next, we will identify the controversial texts, remove their 
annotations and so as to obtain high-quality corpora.

\begin{table}
\caption{The average $\alpha_U$ for students' annotations
  on gold standard texts,
  before and after the removal of less-devoted students' annotations.}
\label{table:alpha_of_golddata}
\centering
\begin{tabular}{l|c|c|c|c|c}
\hline
{\bf } & {\bf gold 1} & {\bf gold 2} & {\bf gold 3} & {\bf gold 4} & {\bf gold 5}\\\hline
Before & 0.4713 & 0.3335 & 0.2146 & 0.1198 & 0.0869 \\\hline
After & 0.6555 & 0.4551 & 0.3645 & 0.1986 & 0.1027 \\\hline
\end{tabular}
\end{table}

\subsection{Dealing With Controversial Annotations and
Obtain the Final Corpora}
\label{subsec:post_process:controversial}

By manually reading and analysing the annotations,
we find that the hotel reviews generally fall into two categories:
i) easy reviews, 
in which the argument component type of each sentence
is quite clear and their relations are easy to identify; and
ii) controversial reviews, in which a high percentage (over 30\%) of
sentences meet multiple argument component types' definitions,
and their labels are heavily dependent on their contexts. 
For the easy reviews, 
we can obtain the annotations for 
the argument component, sentiments and relations;
for the controversial reviews, although many sentences
have controversial annotations, we may still be able to 
find some less-controversial sentences and obtain their annotations.
Thus, we build two corpora based on the annotations
we have collected: one consists of easy reviews,
and the other consists of the relatively 
less-controversial sentences in controversial reviews,
so as to extract as much useful information from the annotations
as possible. 

Reviews whose $\alpha_U$ scores are equal or larger 
than 0.6 are marked as easy 
reviews; we find that the agreement for relations
and sentiments annotations are also high among these reviews
(details are given in Sect. \ref{subsec:post_process:controversial}).
In total, 316 hotel reviews are marked as easy reviews.
The remaining 1911 hotel reviews are marked as controversial,
and we segment them by sentences, so as to find 
the less-controversial sentences therein.
In total, the controversial hotel reviews include
5212 sentences; among them, sentences whose $\alpha_U$
is equal or larger than 0.7 are marked as less-controversial
sentences, accounting for around one-fourth sentences (1452/5212). 

\begin{figure}
\centering
\includegraphics[width=0.48\textwidth]{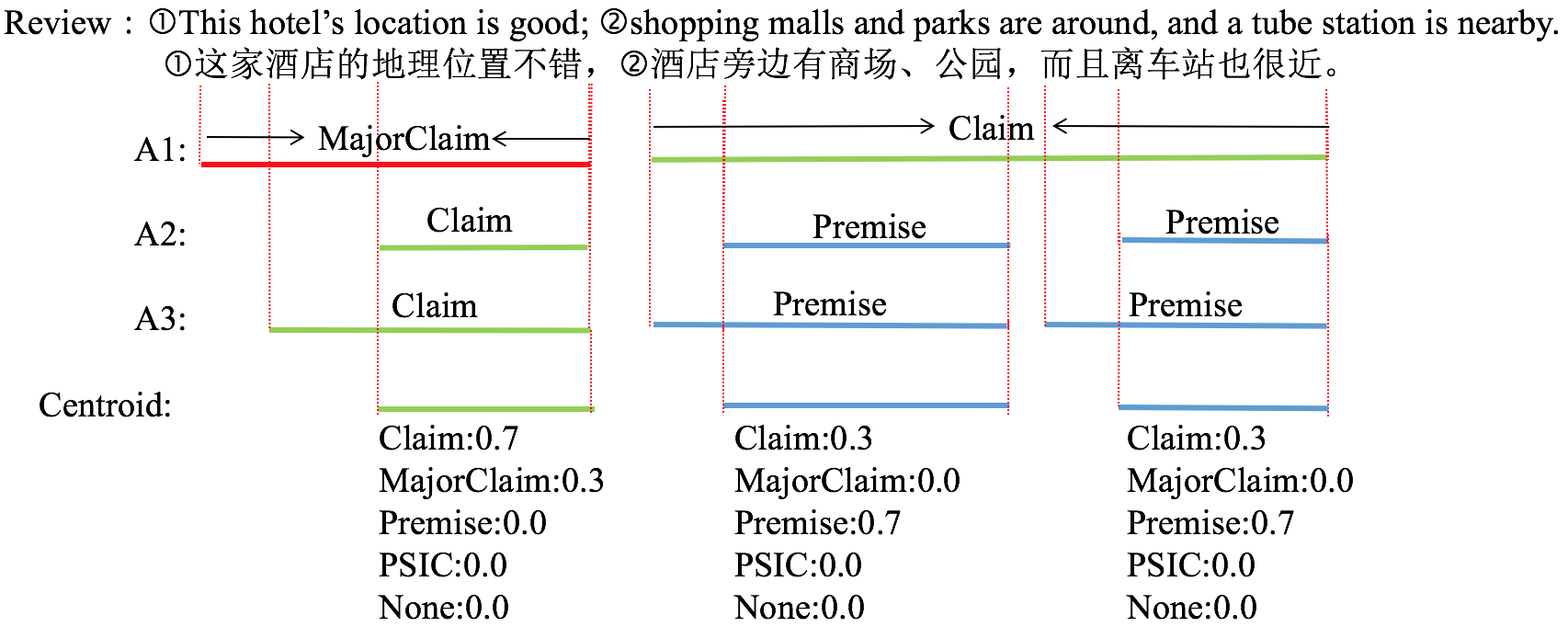}
\caption{An illustrative example of annotations aggregation.}
\label{fig:cluster-example}
\end{figure}

\subsubsection{Easy Reviews Corpus}
\label{subsubsec:easy_review_corpus}
We first need to aggregate the annotations for
argument components, which lays the foundation
for aggregating sentiments and relations.
As a concrete example, consider a
sentence and its annotations presented in Fig. \ref{fig:cluster-example}. 
We can see that three annotations are different in
their boundaries annotations as well as on their argument
component types annotations; to obtain a converged annotation
from these diverging annotations, two sub-tasks
are involved: 
i) resolve the conflicts between argument
component boundaries, so as to obtain
clause texts, and 
ii) decide the argument component type
for the obtained clause.

We employ the K-means clustering technique \cite{hartigan1979algorithm}
to perform the argument component aggregation,
which can perform the above two subtasks
as a whole. Specifically, we vectorise
each student's annotation in one-hot manner: 
each character is represented by 5 digits, 
representing the student's annotation for this 
character. 
We perform 1-cluster clustering on the 
annotation vectors, and the centroid of the clustering 
gives the boundary as well as the argument component type.
Still consider the example in Fig. \ref{fig:cluster-example}:
the first 5 digits in the centroid is (0.33,0,0,0,0.67),
where the first digit corresponds to MajorClaim and 
the last digit corresponds to Non-Argumentative,
and thus we label it as Non-argumentative, and the 
confidence of this label is 0.67.
By computing the centroid, we not only aggregate the component
annotations, but also obtain the confidence score 
for each label. Some statistics of the argument components
annotations in easy reviews 
are presented in Table \ref{table:ira-easy}, and according to our statistics, the average of each review is annotated by three students;
as for IRA metrics, besides $\alpha_U$, we  
also use percentage agreement, multi-$\pi$ \cite{fleiss1971measuring} 
and Krippendorff's $\alpha$ \cite{klaus1980content}.


Based on the converged annotations for components, 
we first evaluate the quality of annotations for sentiments
and reviews in the easy hotel reviews.
Table \ref{table:ira-easy-sentiment-relation} 
presents the agreement scores 
for sentiments and relations.
We can see that almost all scores
are over 0.5, suggesting the agreement is substantial.
Thus, we directly employ the majority voting technique
to aggregate the annotations for relations and sentiments,
and obtain the easy review corpus. 

\begin{table}[!t]
\caption{Some statistics and agreement scores for argument components in 
the easy reviews corpus. There are 316 reviews in this corpus, the column of avg. denotes average per review}
\label{table:ira-easy}
\centering
\begin{tabular}{|l|c|c|c|c|c|c|}
\hline
& \multicolumn{2}{|c|}{Statistic} & \multicolumn{4}{|c|}{IRA} \\\hline
{Label} & {Total} & {avg.}& {\%} & {$\pi$} & {$\alpha$} & {$\alpha_{U}$}\\\hline
Major Claim & 137& 0.4 & 0.928 & 0.516 & 0.542 & 0.430  \\\hline
Claim & 1418 & 4.5& 0.889& 0.743 & 0.752 & 0.694\\\hline
Premise & 689 & 2.2 & 0.873 & 0.683 & 0.696 & 0.612\\\hline
PSIC & 98 & 0.3 & 0.900 & 0.414 & 0.432 & 0.138\\\hline
\end{tabular}
\end{table}

\begin{table}[!t]
\caption{
Some statistics and agreement scores for sentiment and relation in 
the easy reviews corpus.}

\label{table:ira-easy-sentiment-relation}
\centering
\begin{tabular}{|l|c|c|c|c|}
\hline
 & \multicolumn{1}{|c|}{Statistic} & \multicolumn{3}{|c|}{IRA} \\\hline
{Label/Relation} & {Total}&{\%} & {$\pi$} & {$\alpha$}\\\hline
 MC(Positive) & 130& 0.761 & 0.439 & 0.608 \\\hline
 MC(Neutral) & 3 & 0.973 & 0.939 & 0.960 \\\hline
 MC(Negative) & 4 & 0.976 &0.947 & 0.968 \\\hline 
 Claim(Positive) & 1201 & 0.837 & 0.470 & 0.516\\\hline
 Claim(Neutral) & 60 & 0.960 & 0.790 & 0.805 \\\hline
 Claim(Negative) & 157 & 0.961 & 0.845 & 0.855 \\\hline 
 Support(222 reviews) & 399 & 0.918 & 0.702 & 0.716 \\\hline
 Attack(222 reviews) & 12 & 0.996 & 0.908 & 0.911 \\\hline
\end{tabular}
\end{table}

\subsubsection{Less-Controversial Sentences Corpus}
\label{subsubsec:sentence_corpus}

Again, we use K-means to aggregate the
annotations for argument component in 
the less-controversial sentences. 
Some statistics of argument component annotations 
are presented in Table \ref{table:ira-hard}, and each sentence has an average of three students to annotate.
We can see that the size
of the less-controversial sentence corpus is even larger
than the less-controversial review corpus, indicating that
much useful information can be extracted even from the 
highly diverging annotations.
As for the relations annotations in controversial hotel reviews,
among all pairs of less-controversial sentences, 
only 5\% have been annotated (by as least one annotator) as having relations;
thus, we do not aggregate the relation annotations
for the less-controversial sentences. The sentiment 
scores for these sentences are presented in Table 
\ref{table:ira-hard-sentiment}.

\begin{table}[!t]
\caption{
Some statistics and agreement scores for argument components in 
the less-controversial sentences corpus. There are 1452 sentences, the column of avg. denotes average per sentence} 
\label{table:ira-hard}
\centering
\begin{tabular}{|l|c|c|c|c|c|c|}
\hline 
 & \multicolumn{2}{|c|}{Statistic} & \multicolumn{4}{|c|}{IRA} \\\hline
 {Label}& {Total} & {avg.}& {\%} & {$\pi$} & {$\alpha$} & {$\alpha_{U}$}\\\hline
Major Claim & 226 &0.2& 0.915 & 0.728 &0.774 & 0.829 \\\hline
Claim & 1466 & 1.0 & 0.933 &0.817 & 0.844 & 0.918\\\hline
Premise & 653 & 0.4 & 0.888 & 0.688 & 0.735 & 0.779\\\hline
PSIC & 127 & 0.1 & 0.748 & 0.349 &0.453 &0.679 \\\hline
\end{tabular}
\end{table}

\begin{table}[!t]
\caption{
Some statistics and agreement scores for sentiment in 
the less-controversial sentences corpus.
}
\label{table:ira-hard-sentiment}
\centering
\begin{tabular}{|l|c|c|c|c|}
\hline
 & \multicolumn{1}{|c|}{Statistic} & \multicolumn{3}{|c|}{IRA} \\\hline
 {Label}& {Total}&{\%} & {$\pi$} & {$\alpha$}\\\hline
 MC(Positive) & 215& 0.899 & 0.774 & 0.852 \\\hline
 MC(Neutral) & 0 & 0.991 & 0.982 & 0.991 \\\hline
 MC(Negative) & 11 & 0.981 &0.956 & 0.973 \\\hline 
 Claim(Positive) & 1215 & 0.882 & 0.700 & 0.778\\\hline
 Claim(Neutral) & 49 & 0.957 & 0.891 & 0.922 \\\hline
 Claim(Negative) & 198 & 0.962 & 0.908 & 0.934 \\\hline 
\end{tabular}
\end{table}

\subsection{Error Analysis}
\label{subsec:error_analysis}

In order to study the disagreements in the annotations, 
we created confusion probability matrices (CPM) \cite{cinkova2012managing} 
for argument components annotations. 
A CPM contains the conditional probabilities that 
an annotator gives a certain label (column) given 
that another annotator has chosen the label
in the row for a specific item. For example, 
CPM for the easy review corpus and 
the less-controversial sentences corpus are presented 
in Table \ref{table:CPM1},
and the upper-left cell in Table \ref{table:CPM1}
means that, when some other annotators
have labelled a clause as MajorClaim, an annotator will
have 0.481 probability to label this clause also as MajorClaim.

\begin{table}[!t]
\caption{CPM for argument components in the easy reviews 
corpus and the less-controversial sentences corpus.
``MC'' stands for major claim.}
\label{table:CPM1}
\centering
\begin{tabular}{|l|c|c|c|c|c|}
\hline
{Reviews Corpus} & {MC} & {Claim} & {Premise} & {PSIC} & {NA}\\\hline
MC & 0.481 & 0.262 & 0.067 & 0.005 & 0.183 \\\hline
Claim & 0.031 & 0.755 & 0.139 & 0.009 & 0.064 \\\hline
Premise & 0.009 & 0.168 & 0.679 & 0.019 & 0.123 \\\hline
PSIC & 0.009 & 0.126 & 0.209 & 0.453 & 0.200 \\ \hline
NA & 0.058 & 0.173 & 0.273 & 0.416 & 0.453 \\\hline
\hline
{Sentences Corpus} & {MC} & {Claim} & {Premise} & {PSIC} & {NA}\\\hline
MC & 0.800 & 0.088 & 0.026 & 0.002 & 0.082 \\\hline
Claim & 0.012 & 0.877 & 0.074 & 0.002 & 0.035 \\\hline
Premise & 0.006 & 0.118 & 0.781 & 0.010 & 0.086 \\\hline
PSIC & 0.007 & 0.037 & 0.123 & 0.673 & 0.160 \\ \hline
NA & 0.050 & 0.154 & 0.234 & 0.035 & 0.528 \\\hline
\end{tabular}
\end{table}
 
From Table \ref{table:CPM1} we can see that 
there exist no significant
confusion between annotations in the less-controversial
sentences corpus, and we believe the reason is the 
strict criterion for selecting less-controversial 
sentences ($\alpha_U \geq 0.7$; see Sect. 
\ref{subsec:post_process:controversial}). 
However, in the easy reviews corpus (see 
Table \ref{table:CPM1}), we find that 
the confusion between NA and PSIC is significant.
As a concrete example, consider the following sentence
obtained from the easy reviews corpus:
``酒店旁边有个很有特色的酒吧，非常喜欢那里。''
(``There is a very special bar next to the hotel; I like
the bar very much."). 
Some annotators label this sentence as a PSIC, 
as they believe that this sentence supports an 
implicit claim ``hotel locates at a convenient place'';
however, some other annotators label this 
sentence as NA,
because the sentence says nothing about the hotel itself.
As we do not provide a candidate list 
of implicit claims, different annotators
naturally have different understandings of PSIC,
resulting in the high confusion between PSIC and NA.
A possible solution is to provide a candidate
list of implicit claims; we leave this 
as future work.

\section{Conclusion}
\label{sec:conclusion}

In this work, we present the first Chinese argumentation 
corpus, and present the crowdsourcing technique
we used to build this corpus. The argumentation
model used in corpus extends some classic models,
and we believe it is suitable for product reviews
in general. In particular, we novelly use the 
clustering technique to aggregate annotations,
which can not only resolve annotation conflicts, 
but also provide a confidence score at the same time. 
The annotation quality of our corpus is comparable
to some widely used argumentation corpora in other 
languages. To stimulate further research, we make 
the corpus publicly available\footnote{Download website: 124.16.136.215}.


\bibliography{general_short}

\begin{thebibliography}{10}

\bibitem{cinkova2012managing}
Silvie Cinkov{\'a}, Martin Holub, and Vincent Kr{\'\i}{\v{z}}.
\newblock Managing uncertainty in semantic tagging.
\newblock In {\em Proceedings of the 13th Conference of the European Chapter of
  the Association for Computational Linguistics}, pages 840--850. Association
  for Computational Linguistics, 2012.

\bibitem{fleiss1971measuring}
J.~L. Fleiss.
\newblock Measuring nominal scale agreement among many raters.
\newblock {\em Psychological Bulletin}, 76(5):378, 1971.

\bibitem{ghosh2014crowdsourcing}
D.~Ghosh, S.~Muresan, N.~Wacholder, M.~Aakhus, and M.~Mitsui.
\newblock Analyzing argumentative discourse units in online interactions.
\newblock In {\em Proc. of Workshop on Argumentation Mining}, pages 39--48,
  2014.

\bibitem{habernal2014argumentation}
I.~Habernal, J.~Eckle-Kohler, and I.~Gurevych.
\newblock Argumentation mining on the web from information seeking perspective.
\newblock In {\em Proc. of the Workshop on Frontiers and Connections between
  Argumentation Theory and Natural Language Processing}, 2014.

\bibitem{ig2016web}
I.~Habernal and I.~Gurevych.
\newblock Argumentation mining in user-generated web discourse.
\newblock {\em Computational Linguistics}, 2016.

\bibitem{hartigan1979algorithm}
John~A Hartigan and Manchek~A Wong.
\newblock Algorithm as 136: A k-means clustering algorithm.
\newblock {\em Journal of the Royal Statistical Society. Series C (Applied
  Statistics)}, 28(1):100--108, 1979.

\bibitem{kawahara2014crowdsourcing}
D.~Kawahara, Y.~Machida, T.~Shibata, and S.~Kurohashi.
\newblock Rapid development of a corpus with discourse annotations using
  two-stage crowdsourcing.
\newblock In {\em Proc. of COLING}, pages 269--278, 2014.

\bibitem{klaus1980content}
Krippendorff Klaus.
\newblock Content analysis: An introduction to its methodology, 1980.

\bibitem{krippendorff2004measuring}
Klaus Krippendorff.
\newblock Measuring the reliability of qualitative text analysis data.
\newblock {\em Quality \& quantity}, 38(6):787--800, 2004.

\bibitem{lippi2016survey}
M.~Lippi and P.~Torroni.
\newblock Argumentation mining: State of the art and emerging trends.
\newblock {\em ACM Transactions on Internet Technology}, 2015.

\bibitem{moens2013argumentation}
Marie-Francine Moens.
\newblock Argumentation mining: Where are we now, where do we want to be and
  how do we get there?
\newblock In {\em Proc. of Forum on Information Retrieval Evaluation}, 2013.

\bibitem{palau2009argumentation}
R.~M. Palau and M.-F. Moens.
\newblock Argumentation mining: the detection, classification and structure of
  arguments in text.
\newblock In {\em Proc. of ICAIL}, 2009.

\bibitem{reed2008corpus}
C.~Reed, R.~Mochales-Palau, G.~Rowe, and M.-F. Moens.
\newblock Language resources for studying argument.
\newblock In {\em Proc. of LREC}, 2008.

\bibitem{snow2008crowdsourcing}
R.~Snow, B.~O'Connor, D.~Jurafsky, and A.~Y. Ng.
\newblock Cheap and fast—but is it good?: evaluating non-expert annotations
  for natural language tasks.
\newblock In {\em Proc. of EMNLP}, pages 254--263, 2008.

\bibitem{ig2014coling}
C.~Stab and I.~Gurevych.
\newblock Annotating argument components and relations in persuasive essays.
\newblock In {\em Proc. of COLING}, 2014.

\bibitem{ig2016parsing}
C.~Stab and I.~Gurevych.
\newblock Parsing argumentation structures in persuasive essays.
\newblock {\em arXiv preprint}, arXiv:1604.07370, 2016.

\bibitem{stenetorp2012brat}
P.~Stenetorp, S.~Pyysalo, G.~Topic, T.~Ohta, S.~Ananiadou, and J.~Tsujii.
\newblock brat: a web-based tool for nlp-assisted text annotation.
\newblock In {\em Proc. of ECAL}, pages 102--107, 2012.

\bibitem{villalba2012some}
M.~P.~Garcia Villalba and P.~Saint-Dizier.
\newblock Some facets of argument mining for opinion analysis.
\newblock In {\em Proc. of COMMA}, 2012.

\bibitem{wachsmuth2014review}
H.~Wachsmuth, M.~Trenkmann, B.~Stein, G.~Engels, and T.~Palakarska.
\newblock A review corpus for argumentation analysis.
\newblock In {\em Computational Linguistics and Intelligent Text Processing},
  pages 115--127. 2014.

\bibitem{wyner2012semi}
A.~Wyner, J.~Schneider, K.~Atkinson, and T.~JM Bench-Capon.
\newblock Semi-automated argumentative analysis of online product reviews.
\newblock In {\em Proc. of COMMA}, 2012.

\end{thebibliography}
\bibliographystyle{plain}

\end{CJK*}

\end{document}